\documentclass[11pt,a4paper]{article}
\usepackage[hyperref]{acl2019}
\usepackage{url}
\usepackage{times}
\usepackage{latexsym}
\usepackage{booktabs}
\usepackage{arabtex}
\usepackage{multirow}
\usepackage{array}
\usepackage{enumitem}
\usepackage{amsmath, amsthm}
\usepackage{amssymb}
\usepackage{xfrac}
\usepackage{utf8}
\usepackage{changepage}
\setcode{utf8}
\usepackage{comment}
\aclfinalcopy 

\title{Neural Arabic Question Answering}

\author{ Hussein Mozannar, Karl El Hajal, Elie Maamary, Hazem Hajj \\
  Department of Electrical and Computer Engineering \\
  American University of Beirut \\
  {\{hssein.mzannar, karlhajal, eliemaamary17\}@gmail.com, hh63@aub.edu.lb}
  }
\date{}

\begin{document}
\maketitle


\centerline{\large\bf Abstract}
\begin{adjustwidth}{0.6cm}{0.6cm}
\fontsize{10}{12}\selectfont
    This paper tackles the problem of open domain factual Arabic question answering (QA) using Wikipedia as our knowledge source. This constrains the answer of any question to be a span of text in Wikipedia. Open domain QA for Arabic entails three challenges: annotated QA datasets in Arabic, large scale efficient information retrieval and machine reading comprehension. To deal with the lack of Arabic QA datasets we present the Arabic Reading Comprehension Dataset (ARCD) composed of 1,395 questions posed by crowdworkers on Wikipedia articles, and a machine translation of the Stanford Question Answering Dataset (Arabic-SQuAD). Our system for open domain question answering in Arabic (SOQAL) is based on two components: (1) a document retriever using a hierarchical TF-IDF approach and (2) a neural reading comprehension model using the pre-trained bi-directional transformer BERT.  Our experiments on ARCD indicate the effectiveness of our approach with our BERT-based reader achieving a 61.3 F1 score, and our open domain system SOQAL achieving a 27.6 F1 score.
\end{adjustwidth}

\section{Introduction}
One of the goals in artificial intelligence (AI) is to build automated systems that can perform open-domain question answering (QA) through understanding natural language and gathering knowledge \cite{googlenaturalquestion}.
The driver behind progress in English QA has been the release of massive datasets including the Stanford Question Answering Dataset (SQuAD), WikiQA \cite{rajpurkar2016squad,yang2015wikiqa}. The task in these datasets is to find the span of text in a document that answers a given question. 
On the other hand, progress in Arabic QA systems has lagged behind their English counterparts. While there has been a good body of work on methods for question answering, they mostly have a common limitation of being tested on small amounts of data and relying on classical methods \cite{shaheen2014arabic}.

In this work, we tackle the problem of answering Arabic open-domain factual questions using Arabic Wikipedia as our knowledge source. The open-domain setting poses many challenges, from efficient large scale information retrieval, to highly accurate answer extraction modules, and this requires a sizable amount of data for training and testing.

\begin{figure}[t!]
\centering
\resizebox{7.7cm}{!}{%
\begin{tabular}{p{7.7cm}}
\hline

\begin{flushright}
\small{
\<  
يلعب نادي نايفربول كل مبارياته الرسمية في \textbf{ ملعب الأنفيلد }، \break و الذي يتسع لحضور 54,074 متفرج.
 \break
 يعتبر نادي مانشستر يونايتد العدو اللدود لنادي
ليفربول،  فتلك\break  المواجهات تعتبر من أهم المنافسات في كرة  
القدم الإنجليزية،\break  حيث تجمع أكثر ناديي تحقيقاً للألقاب،
حيث حقق مانشستر\break  يونايتد 62، بينما حقق ليفربول \textbf{59 بطولة }.
>
}
\end{flushright}
\\
\begin{flushright}
\<
كم من بطولة حققها نادي ليفربول؟
>
\end{flushright}
\begin{flushright}
\<
\textbf{
59 بطولة
}
>
\end{flushright}
\\
\begin{flushright}
\<
أين يلعب نادي ليفربول مبارياته ؟
>
\end{flushright}
\begin{flushright}
\<
\textbf{
ملعب الأنفيلد
}
>
\end{flushright}
\\
\hline
\end{tabular}%
}
\caption{Example data point from ARCD containing a paragraph with two accompanying questions}
\label{arcd:example}
\end{figure}

First, to deal with the need of large Arabic reading comprehension datasets, we develop the following: (1) The Arabic Reading Comprehension Dataset (ARCD) composed of 1,395 crowdsourced questions with accompanying text segments on Arabic Wikipedia as seen in figure \ref{arcd:example}, and (2) Arabic-SQuAD  consisting of 48k paragraph-question-answer machine translated tuples from the SQuAD dataset.

Second, modern open-domain QA systems are generally composed of two parts: a retriever that obtains relevant segments of text, and a machine reading comprehension (MRC) model that extracts the answer from the text \cite{chen2017reading}. For our retriever, we propose the use of a hierarchical TF-IDF retriever that is efficiently able to trade off between n-gram features and the number of documents retrieved. 
We chose raw Wikipedia text as our information source instead of knowledge bases \cite{lehmann2015dbpedia} which are commonly used for open-ended QA as it enables our approach to tackle other domains and settings with little adaptation.
Now there has been remarkable progress in designing neural MRC models that read and extract answers from short paragraphs; we selected two of the best performing models on the SQuAD dataset \cite{rajpurkar2016squad} as our document readers. The first is QANet \cite{yu2018qanet}, an efficient convolution and self-attention-based neural network, and the second is BERT \cite{devlin2018bert}, a transformer-based pre-trained model. From the document retriever and reader we build an open domain QA system named SOQAL by combining confidence scores from each.


We evaluated our system components on the crowdsoured ARCD dataset: Our hierarchical TF-IDF retriever is competitive with Google Search, and our BERT reader is the current state-of-the-art for reading comprehension. Finally, our open domain system SOQAL achieves a respectable 27.6 F1 on ARCD.   

To summarize, the contributions of the paper are:
\begin{itemize}[noitemsep]
    \item  \textbf{Datasets for Arabic QA.} Crowdsourced Arabic Reading Comprehension Dataset (ARCD) of 1,395 questions, and translated Arabic-SQuAD: 48k translated questions from \cite{rajpurkar2016squad}.
    \item \textbf{Neural Reading comprehension in Arabic.} State of the art MRC models for Arabic based on BERT \cite{devlin2018bert} and QANet \cite{yu2018qanet}.
    \item \textbf{Open domain Arabic QA system.} End-to-end system for open domain Arabic questions using a hierarchical TF-IDF retriever, BERT and linear answer ranking. 
\end{itemize}

All the data and system implementation is available at \url{https://github.com/husseinmozannar/SOQAL}.
\section{Related Work}
\textbf{Open-domain Arabic question answering.} The state of current Arabic QA systems is summarized in \cite{shaheen2014arabic}: research has focused mostly on open-ended QA using classical information retrieval (IR) methods, and there are no common datasets for comparisons. Consequently, progress has been slow. 
Furthermore, the Arabic language presents its own set of difficulties: given the highly intricate nature of the language, proper understanding can be difficult. For instance, \<فسيأكلونه> means ``so they will eat it", which demonstrates the complexity that can be presented by a single word. Moreover, Arabic words require diacritization for their meaning to be completely understood. For example, \<عَلَّمَ> translates into ``he taught", and \< عَلِمَ> means ``found out"; modifying one diacritic changes the meaning entirely.

\begin{table}
\centering
\resizebox{7.7cm}{!}{%
\begin{tabular}{p{4.1cm}p{2.3cm}p{2.0cm}r}
 \toprule
 \textbf{Dataset} & \textbf{Source} & \textbf{Formulation} &\textbf{Size}\\
 \midrule
\textbf{Arabic-SQuAD }& \textbf{Translated SQuAD} & \textbf{p,q,a} & \textbf{48,344} \\
\textbf{ARCD} & \textbf{Arabic Wikipedia} & \textbf{p,q,a}& \textbf{1,395} \\
\hline
ArabiQA \break \small{\citep{arabiqa}} & Wikipedia & q,a& 200 \\
DefArabicQA \break \small{\citep{trigui2010defarabicqa}} & Wikipedia and Google search engine  & q,a with documents& 50 \\
\flushleft{Translated TREC and CLEF}  \break \small{\citep{abounour}} & \flushleft{Translated TREC and CLEF}   & q,a& 2,264 \\
QAM4MRE  \break \small{\citep{QA4MRE}} & selected topics   & document,q and multiple answers& 160 \\
DAWQUAS  \break \small{\citep{DAWQAS}} & auto-generated from web scrape& q,a &3205 \\
QArabPro  \break \small{\citep{akour2011qarabpro}} & Wikipedia & q,a &335 \\
\bottomrule
\end{tabular}%
}
\caption{Available question answering datasets in Arabic. p:paragraph, q:question and a:answer  }
\label{table:datasets}
\end{table}

We now review some of the methods and datasets used in the literature and compare them in table \ref{table:datasets}.
Most of the datasets listed are of very limited size and do not include accompanying text segments so as to enable reading comprehension. Furthermore, all datasets with size bigger than 1000 questions are synthetically generated. 
 Approaches have tackled specific types of questions and are heavily dependent on their nature focusing more on document retreival. In \cite{azmi2016answering}, they attempt to answer "why" questions using classic IR methods and rhetorical structure theory, and their methods are evaluated on a set of 100 questions. 
 On the other hand, DefArabicQA \cite{trigui2010defarabicqa} focuses on definition question and uses an answer ranking module based on word frequency. 
 QArabPro \cite{akour2011qarabpro} employs a rule-based question answering system and obtains an 84\% accuracy on 335 questions based on Wikipedia. The SemEval task 3 in 2015, 2016, and 2017 \cite{nakov2017semeval} tackled community question answering. It included a task in Arabic with each data point consisting of a paragraph, a question, and multiple answers, and the goal was to rank them in order of relevance.
 One of the strategies used to solve the 2015 edition  was to train an SVM ranker by embedding the questions and answers using Word2vec \cite{belinkov2015answer}. The type of data used is not constructive for training answer extraction systems but can be helpful for recognizing relevance.

\textbf{QA Datasets.} 
As previously mentioned, the driver behind progress in QA has been the release of large datasets in addition to advances in deep learning and language representation models \cite{devlin2018bert}.
The most popular benchmark for reading comprehension has been the Stanford Question Answering Dataset \cite{rajpurkar2016squad}. Other notable datasets include: WikiQA \cite{yang2015wikiqa}, a sentence selection task using Wikipedia passages, and TriviaQA  \cite{joshi2017triviaqa}, a dataset of trivia questions with provided evidence.

\textbf{Reading comprehension and  QA. }
Recently, machine reading comprehension has made significant progress using recurrent models and attention mechanisms to capture long term interactions \cite{seo2016bidirectional}, and this has prompted its use as part of open-domain QA. On the other hand, given that recurrent networks are slow in training and inference, QANet \cite{yu2018qanet} proposes an approach based only on convolutions and self-attention that is able to achieve very competitive results on SQuAD while being 10x faster than recurrent based approaches such as Bidirectional Attention Flow (BiDAF) \cite{seo2016bidirectional}. For open-domain QA, \cite{chen2017reading} investigates the use of Wikipedia as a knowledge source and implements a two component system based on a TF-IDF retriever and a RNN reader achieving a 29.8\% exact- match accuracy on open-SQuAD. 
 Other approaches have attempted to build more sophisticated retrievers by formulating it as a reinforcement learning problem \cite{wang2018joint,wang2018r}, or as a supervised learning problem using distant supervision for data \cite{das2018multi,lin2018denoising}.


In the following sections we will first describe the datasets collected, and then our proposed method for Arabic open-domain question answering.

\section{Dataset Collection}
\subsection{Arabic Reading Comprehension Dataset}
To properly evaluate our system, we must have  questions written by proficient Arabic speakers, and thus we resort to crowdsourcing to develop our dataset.
\begin{figure}
    \centering
    \includegraphics[width=7.8cm,height=8.5cm]{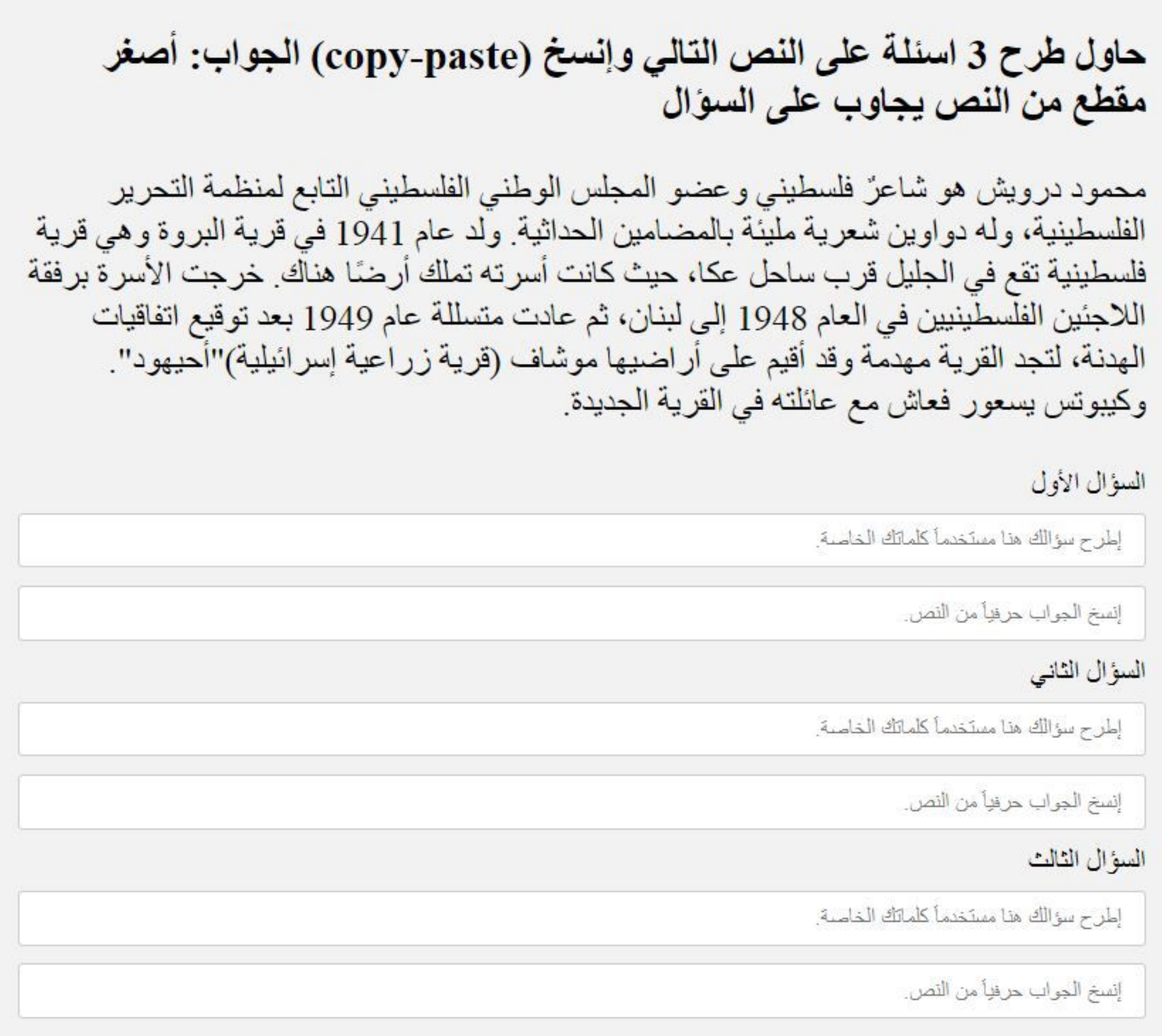}
    \caption{Interface for the crowdworkers}
    \label{fig:interface}
\end{figure}

\textbf{Task Description.} Each task presented to the crowdworkers consists of five articles taken from Arabic Wikipedia, from which we extracted the first three paragraphs with a length greater than 250 characters. The worker has to write three question-answer pairs for each paragraph in clear Modern Standard Arabic, where the answer to each question should be an exact span of text from the paragraph. The interface, shown in figure \ref{fig:interface},
consists of a paragraph along with two text boxes for each of the 3 question-answer pairs. Pasting is disabled in the question fields in order to encourage workers to use their own words, but it is enforced in the answer fields to guarantee that the answer is taken as-is from the paragraph. Before workers begin the task, they have to answer a reading comprehension question from a test set we created to make sure of their language proficiency. Only workers who succeeded in the test were accepted.

\textbf{Article curation.} The articles presented in the tasks were 155 articles randomly sampled from the 1000 most viewed articles on Wikipedia in 2018. We used MediaWiki's API\footnote{Availabe at \url{https://en.wikipedia.org/w/api.php}
} to retrieve the most viewed articles per month in 2018 for Arabic Wikipedia and aggregated the results. The articles covered a diverse set of topics including religious and historical figures, sports celebrities, countries, and companies. We additionally manually filtered out adult content.

\textbf{Crowdsourcing.} We resorted to Amazon Mechanical Turk for crowdsourcing.
Crowdworkers were required to have a minimum HIT acceptance of 97\%, and at least 100 HITs submitted.
Moreover, our task description highlighted the need for good Arabic skills. Workers were advised to spend 3 to 4 minutes per paragraph and were paid close to 10 USD per hour. They were encouraged to ask difficult questions framed in such a way that they can be answered outside the scope of the paragraph.
In total, we collected 1,395 questions based on 465 paragraphs from 155 articles based 
on the Amazon Turk HITs.

\subsection{Arabic-SQuAD}

\textbf{Translating SQuAD.} While the crowdsourcing of questions by proficient Arabic writers is essential to properly evaluate our systems, noisy data could well suffice for training. Indeed, backtranslation as a means for data augmentation has been effective in improving the performance of neural MRC \cite{yu2018qanet}, and this gives hope that translated data could be used to train our machine reading comprehension module. We chose to translate SQuAD version 1.1 \cite{rajpurkar2016squad}. It is currently the most popular benchmark for MRC and was collected through crowdsourcing based on Wikipedia articles. SQuAD contains 107,785 paragraph-question-answer tuples on 536 articles, and we translated the first 231 articles of the SQuAD training set using the Google Translate neural machine translation (NMT) API \cite{wu2016google}. This resulted in 48,344 questions on 10,364 paragraphs.
\begin{figure*}[h]
    \centering
    \includegraphics[trim={0.6cm 0cm 0cm 2cm},clip,scale=0.4]{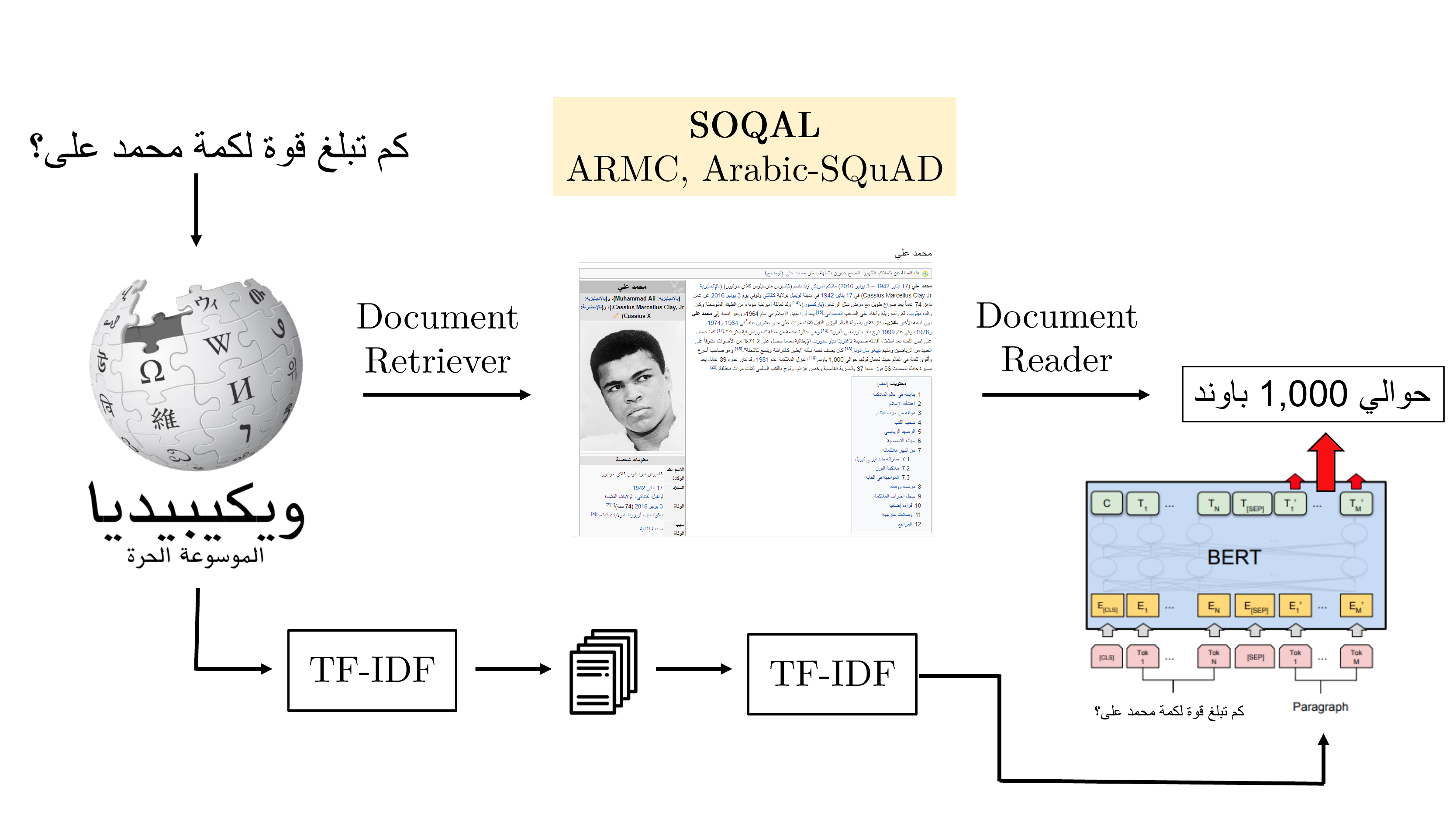}
    \caption{Architecture of our open domain question answering system SOQAL. BERT illustration is adapted from \cite{devlin2018bert}}
    \label{fig:system}
\end{figure*}
\section{ Our System: SOQAL}

We will now describe the architecture of our system for open domain question answering for the Arabic language (SOQAL). It is composed of three modules: (1) a document retriever that obtains relevant documents to the question, (2) a machine reading comprehension module that extracts answers from the documents retrieved, and an (3) answer ranking module that ranks the answers in order of relevance by taking in scores from both the document retriever and the reader. The inputs to the system are a question consisting of $m$ tokens $q=\{q_1,\cdots,q_m\}$, and the entirety of Arabic Wikipedia, and its output is a small span of text extracted from Wikipedia which should answer the question. The pipeline is illustrated in figure \ref{fig:system}.

\subsection{ Hierarchical TF-IDF Document Retriever}
The goal of this module is to select the documents that are most relevant to the question, thus reducing the span of search of our reader.
Arabic Wikipedia is made up of 664,768 indexed articles with an average of 3.4 paragraphs per article, totalling 2,683,743 paragraphs with an average of 233 characters per paragraph. We discard imagery, lists, and other structured information so that our approach could translate well to various knowledge sources. 

There are two scopes on which we can search: either articles or paragraphs. We denote the set of documents searched over as $D=\{d_1,\cdots,d_n\}$, where for $1\leq i\leq n$, $d_i$ is a single document which can be either an article or a paragraph from an article.



Inspired by classical QA systems \cite{chen2017reading}, we employ a term frequency-inverse document frequency (TF-IDF) based document retriever given its efficiency. Each document is first tokenized and stemmed using the NLTK \cite{bird2006nltk} Arabic tokenizer where stopwords are removed. The TF-IDF matrix of weights of the document set, i.e. Arabic Wikipedia, is then constructed using $n$-gram counts to take into account local word order. As $n$ increases, the retriever becomes more accurate, but the retrieval process becomes slower and more memory prohibitive. 
Each document's vector is normalized. Next, the TF-IDF vector weights of the question are computed based on the vocabulary of the document set. The score for each document is then computed as the cosine similarity between the question and the document vectors. We use a sparse matrix representation for the TF-IDF matrix to speed up computations.
Finally, we return the top $k$ documents with the highest similarity where $k \in \mathbb{N}$ is a hyperparameter. The higher $k$ is, the more likely it is that the set of retrieved documents contains relevant documents, and the slower and more error-prone is the answer extraction process. 

To obtain the benefits of using large $n$-gram features while keeping $k$ small and being computationally efficient, we propose the following hierarchical TF-IDF retriever approach. The first step is to build a TF-IDF retriever on Arabic Wikipedia with bigram features and a very large $k$, say $\approx 1000$, and obtain the set of retrieved documents for a given question, call it $D'$. Then, for each question, we construct a seperate TF-IDF retriever using as document set $D'$ with $4$-gram features and a small $k$, say $\approx 15$. The second retriever does not sacrifice much in terms of the accuracy of the first retrieval step, as $4$-gram features are highly informative and do not add significant computations.
\subsection{BERT Document Reader}
Our proposed reader is Bert \cite{devlin2018bert}, a pre-trained language model that is currently the state of the art on the SQuAD leaderboard \footnote{SQuAD leaderboard \url{https://rajpurkar.github.io/SQuAD-explorer/}}.

Its core model is a bi-directional Transformer \cite{vaswani2017attention}. The input  text is first tokenized using a shared Wordpiece \cite{wu2016google} vocabulary of 104 languages, and it is then embedded; note that Arabic diacritics are removed.
 Each input point of question and paragraph pairs is represented as a single sentence separated by a special token. We need to learn two new vectors: start and end $S,E\in \mathbb{R}^H$ vectors indicating the position of the answer; $H$ is the dimension of the last hidden layer outputs. For each token $i$ in the paragraph, we take the final hidden state of the Transformer $T_i$ and let the probability that $i$ is the start or end of the answer be:
\begin{align*}
&P_{start}(i) \propto \exp(S^T T_i)\\
&P_{end}(i) \propto \exp(E^T T_i)
\end{align*}
Note that we take the un-normalized  exponential to be able to compare across documents.
At inference time we predict the span $(i,j)$ such that $i\leq j\leq i +15$ that maximizes $P_{start}(i)P_{end}(j)$. The training objective is the sum of the log likelihood for each of the start and end positions.
\subsection{Answer Ranking} 

Let us recall the operation of the end-to-end system. The question is first passed to the retriever and the top $k$ documents are gathered; if a document unit is an article then we gather all of its paragraphs. Along with the documents' text, we obtain a score for each document denoted $DocScore(i)$ from the retriever; paragraphs have the same score as their document. For our hierarchical TF-IDF retriever, the scores are the cosine similarities between the document and the question.

The paragraphs obtained from the retriever are each then fed as input to the document reader to obtain candidate answers. We obtain a score for each candidate answer $i$ denoted: $$AnsScore(i) \propto P_{start}(i) \cdot P_{end}(i)$$

 To make sure the answer and document scores are on the same scale, we normalize both individually by passing each through a softmax function. The final step  to obtain the answer of the question is by combining the scores through a linear combination and pick the maximizing answer as follows:
\[
\arg\max_{i \in [k]} \ \ \beta \cdot DocScore(i) + (1-\beta) \cdot AnsScore(i)
\]

Where $\beta \in [0,1]$ is a hyperparameter chosen through a line search using a development set.

As a note, since articles can be very large, one can additionally use a TF-IDF retriever with $4-$ gram features to obtain a smaller set of paragraphs, thus reducing the load on the reader. While this step was not performed for our experimental evaluation, it is crucial when deploying the QA system for usage.

\section{Dataset Analysis}
\subsection{ARCD}
In this section we analyze the properties of the Arabic Reading Comprehension Dataset. To better understand the difficulty of answering the questions, we randomly sampled $100$ questions for the following analysis.

\textbf{Answer diversity.} 
 We, the authors, manually categorized the answers by first separating the numerical and non-numerical answers. Numerical answers were either identified as dates by looking at the question, or were otherwise labeled as other numeric. For the non-numerical answers, we identify the type of phrase as either a verb, adjective, or noun phrase. If it is a noun phrase, we check using MADAMIRA \cite{pasha2014madamira} for named entities, and then manually verify the outcome. The results are shown in table \ref{table:answers}.

\begin{table}
\centering
\resizebox{7.7cm}{!}{%
\begin{tabular}{l l r}
 \hline
 \textbf{Answer type} & \textbf{Percentage} & \textbf{Example} \\ \toprule
Date & \textbf{17\%} & \<10 مارس 1976> \\
Person & \textbf{17\%} & \<الطبيب الشاعر سليم الضاهر> \\
Location & \textbf{10\%} & \<آسيا> \\
Organization & \textbf{9\%} & \<الاتحاد الإنجليزي لكرة القدم>\\
Verb Phrase & \textbf{7\%} & \<انقسمت الإمبراطورية> \\
Adjective Phrase & \textbf{4\%} & \<أقصى اتساع لها> \\
Noun Phrase & \textbf{12\%} & \<الوارد المنحدر>\\
Other Numeric & \textbf{15\%} & \<250 كيلوغرام> \\
Other Entity & \textbf{9\%} & \<جائزة نوبل في الأدب> \\
\bottomrule
\end{tabular}%
}
\caption{ Answer categories percentages in ARCD according to the categorization by \cite{rajpurkar2016squad} }
\label{table:answers}
\end{table}

\textbf{Question Reasoning}
To better understand the reasoning required to answer the questions, we manually labeled the questions according to the following reasoning categories as in \cite{trischler2017newsqa,rajpurkar2016squad}:
\begin{table*}[t!]
\centering
\resizebox{15.4cm}{!}{%
\begin{tabular}{p{3cm}p{11cm}r}
\toprule
  Reasoning & Example & Percentage  \\
  \midrule
  Word matching \break (synonyms) &
  \raggedleft \small{ \< يلعب نادي ليفربول كل مبارياته الرسمية في ملعب الأنفيلد.\break يعتبر نادي مانشستر يونايتد العدو اللدود لنادي ليفربول ، حيث حقق مانشستر \break يونايتد 62، بينما حقق ليفربول\textbf{ 59 بطولة}.  >} \break
 \<
 كم من بطولة حققها نادي ليفربول؟
 > \underline{\ :Q} & \textbf{59\%}\\
  Word matching \break (world knowledge) & 
  \raggedleft \small{ \< نجيب محفوظ (11 ديسمبر 1911 - 30 أغسطس 2006) روائي مصري، \break هو أول عربي حائز على \textbf{جائزة نوبل }في الأدب.  كتب نجيب محفوظ منذ \break بداية الأربعينيات واستمر حتى 2004.   >} \break
 \<
ما هي أهم جائزة عالمية حصل عليها نجيب محفوظ؟
 > \underline{\ :Q}
  &\textbf{15\%}\\   Syntactic variation & \raggedleft \small{ \<   طرح عمر لطفي بك فكرة تأسيس النادي الأهلي في العقد الأول من القرن، \break   لأنه اعتبر أن تأسيس نادي طلبة المدارس العليا سياسيًا بالدرجة الأولى، ووجد \break \textbf{أن هؤلاء الطلبة بحاجة إلى نادٍ رياضي يجمعهم لقضاء وقت الفراغ وممارسة الرياضة}.  >} \break
 \<
لماذا أسس النادي للطلبة؟
 > \underline{\ :Q}
  &\textbf{13\%}\\
  
  Multiple sentence reasoning & \raggedleft \small{ \< سليمان خان الأول بن سليم خان الأول ، عاشر السلاطين العثمانيين وخليفة المسلمين \break الثمانون،  بلغت الدولة الإسلامية في عهده \textbf{أقصى اتساع لها} حتى أصبحت أقوى  \break دولة في العالم في ذلك الوقت.    >} \break
 \<
ماذا بلغت دولة سليمان خان تحت عهده؟
 > \underline{\ :Q} &\textbf{10\%}\\
  Ambiguous &  \raggedleft \small{\<
  متى رسمها؟
  >} \underline{\ :Q} &\textbf{3\%}\\
    \bottomrule
\end{tabular}%
}
\caption{Examples of questions with their respective paragraph (trimmed to fit) and answer in bold from ARCD and the reasoning required to answer them.
  }
  \label{table:questions}
\end{table*}

\begin{itemize}[noitemsep]
    \item \textit{Word matching (synonyms)}: question matches the same word pattern up to synonyms in the paragraph; simple pattern matching is required. 
    \item \textit{Word matching (world knowledge)}: question matches the pattern of the paragraph, however additional inference using world knowledge is required to answer.
    \item \textit{Syntactic variation}:
    The question's syntactic dependency  structure does not match that of the answer sentence.
    \item \textit{Multiple sentence reasoning}: The question draws on knowledge from multiple sentences. Only after making necessary links across sentences can it be answered.
    \item  \textit{Ambiguous}: The question cannot be answered given the information in the paragraph or is unclear.
\end{itemize}

The results and examples are shown in table \ref{table:questions}.

\subsection{Arabic-SQuAD}
We discuss some of the issues resulting from the machine translation of SQuAD and how we handled them.

 We observed that translation performed  well for paragraphs and questions and maintained their original meaning.
The problem is, NMT is heavily context dependent, 
thus identical words and phrases have different translations if the context is varied.
This led to an inconsistency between the translation of the answers and  paragraphs with 25,490 answers not found in their respective paragraphs, almost 47.3\% of the total questions. We remarked that the type of errors that caused the answers to not match in the paragraph mostly arised from two factors: (1) translation was unable to recognize named entities without context and thus transliterated them, and (2) minor typographic like errors from missing or added \<لام التعريف  > (the)
and differing tenses. To fix this issue, we transliterated all the paragraphs and answers to Arabic and found the span of text of length at most 15 words with the least edit-distance with respect to the answer. To verify the efficacy of this approach, we randomly sampled 100 questions where the answer is not found in the paragraph and provided the correct answer. On this test set, the approach managed to exactly find 44\% of the answers, and 64\% of the proposed answers contained the correct answer and did not exceed more than twice its length.

\section{System Experiments}
We now showcase experiments for every component in our system and the end-to-end open domain system.

\textbf{Datasets.} Arabic-SQuAD is split 80-10-10\% into three parts for training, development and testing: Arabic-SQuad-Test is composed of 2,966 questions on 24 articles; note that articles are distinct between the parts. Similarly, ARCD is split 50-50 into training and testing with ARCD-Test having 702 questions on 78 articles.

\subsection{Retriever}
\begin{table}[h]
\centering
\resizebox{7.7cm}{!}{%
\begin{tabular}{l l l}
 \hline
 \textbf{Method} & \textbf{$k$} & \textbf{ARCD}  \\ \hline
Wikipedia API & 15 & 34.8\% \\
Google Search & 10 & \textbf{75.6}\% \\
\hline
TF-IDF Unigram Article & 15 & 41.7\% \\
TF-IDF Bigram Article & 15 & 47.7\% \\
TF-IDF Bigram Article & 350 & 73.5\% \\
Hierarchical TF-IDF & 15 & \textbf{65.3}\% \\
Embedding fastText Paragraph & 50  & 27.0\%\\
\hline
\end{tabular}%
}
\caption{ Comparison of the different retrievers on ARCD. $k$: number of documents retrieved }
\label{table:retriever}
\end{table}

\begin{table*}[h]
\centering
\begin{tabular}{l|c@{\,\,\,\,\,\,}cc|c@{\,\,\,\,\,\,}cc}
\hline
 \bf Method &  \multicolumn{3}{c}{\bf \underline{Arabic-SQuAD Test}} & \multicolumn{3}{c}{\bf \underline{ARCD}} \\
&  EM & F1 & SM & EM & F1 &SM \\ \hline
Random Guess & 0.23 & 4.34 & 23.5 & 0.07 & 8.13 & 51.0 \\
Sliding Win. + Dist. \small{\cite{richardson2013mctest}} & 0.00 & 5.80 & 29.2 & 0.07 & 14.2 & 58.4 \\
Embedding fastText & 0.04 & 6.96 & 43.1 & 0.36 & 15.3 & 73.1  \\
TF-IDF Reader & 0.27 & 2.41  & 49.2 & 0.22 & 5.6 & 75.3  \\
\hline
QANet fastText \small{\cite{yu2018qanet}} & 29.4& 44.4 & 61.7 & 11.0& 38.6 & 83.2 \\
BERT \small{\cite{devlin2018bert}} & \textbf{34.1 } &\textbf{48.6 } & \textbf{66.8 }&\textbf{ 19.6}& \textbf{51.3} & \textbf{91.4} \\
\hline
\end{tabular}
\caption{ Comparison of the different document reader modules on Arabic-SQuAD test set and all of ARCD. QANet and BERT were trained only on the training set of Arabic-SQuAD.}
\label{table:reader}
\end{table*}

We examine the performance of our different
retriever modules on the full ARCD dataset. To compare the approaches we assign to each the ratio of questions for which the answer appears in any of the retrieved document over the total number of questions.

\textbf{Baselines.} We implement three baselines: the first is using Wikipedia's Search API \footnote{\url{https://www.mediawiki.org/wiki/API:Search}}, and the second is through Google Custom Search engine  \footnote{We use the official API \url{ https://developers.google.com/custom-search/}} restricted to the Arabic Wikipedia site. Furthermore, we implement an embedding based retriever using fastText embeddings 300 dimensional Wikipedia pre-trained word embeddings \cite{joulin2016fasttext} that computes for each paragraph a representation using the sum of its word embeddings. Other embedding models exist for Arabic but fastText is the most specialized to Wikipedia \cite{badaro2018ema, al2015deep}

\textbf{Results and Analysis} Our results are reported in table \ref{table:retriever}. We find that even the simple TF-IDF unigram retriever is able to beat the Wikipedia API baseline. Google Search with $k=10$ is the golden standard with 75.6\%, TF-IDF using bigram features and $k=350$ is able to come close with 73.5\%. Using our hierarchical approach of adding a second $4$-gram TF-IDF retriever to a bigram $k=1000$ retriever achieves a respectable 65.3\% improving on the single bigram by 17.6\% and a reduction of 8.2\% from the full $k=350$ retriever. The embedding retriever using fastText \cite{joulin2016fasttext} performed badly in accordance with the results in \cite{chen2017reading}.

It is important to note that since the questions in ARCD were written with a specific paragraph in mind, they might be ambiguous without their context, hence why it is hard to beat the Google Search baseline.

\subsection{Reader}
\begin{table}[h]
\centering
\resizebox{7.7cm}{!}{%
\begin{tabular}{l|c@{\,\,\,\,\,\,}cc}
\hline
 \bf Method &  \multicolumn{3}{c}{\bf \underline{ARCD-Test}} \\
&  EM & F1 & SM \\ \hline
\textbf{Reader:}\\
BERT (SQuAD) & 23.8 & 53.0 & \textbf{90.6 } \\
BERT (ARCD) & 23.9 & 50.1 & 88.0  \\
BERT (SQuAD + ARCD) & \textbf{34.2} & \textbf{61.3} & 90.0  \\
\hline
\textbf{Open-Domain:}\\
SOQAL (top-1) & 12.8& 27.6 & 29.8  \\
SOQAL (top-3) & 17.8 & 37.9 & 44.0  \\
SOQAL (top-5) & 20.7& 42.5 & 51.7  \\
\hline
\end{tabular}%
}
\caption{Results of BERT as a document reader on ARCD-Test under different data regimes and of our open domain system SOQAL when returning the top k answers}
\label{table:opensystem}
\end{table}

\textbf{Metrics.} We evaluate our different readers based on three metrics. The first is
\textit{exact match }(EM) which measures the percentage
of predictions that match the ground truth answer exactly, the second is a
\textit{(macro-averaged) F1 score} \cite{rajpurkar2016squad} that measures
the average overlap between the prediction tokens and the
ground truth answer tokens. Finally, we use a \textit{sentence match} (SM) metric that measures the percentage of predictions that fall in the same sentence in the paragraph as the ground truth answer.

\textbf{Baselines.} We compare against three non-learning baselines. For all three methods, we generate candidate answers by considering every text span of length maximally 10 words in each sentence as a candidate. We implement the following baselines: the sliding window distance based algorithm of \cite{richardson2013mctest}, a TF-IDF reader based on $4$-gram features which operates exactly like the retriever with $k=1$, and finally an embedding  approach where the candidate with the highest cosine similarity with respect to fastText embeddings is returned \cite{joulin2016fasttext,belinkov2015answer}. We also compare against QANet \cite{yu2018qanet}, a competitive MRC network that is especially fast for prediction.

\textbf{Implementation Details.} For Bert, we follow the reference implementation for training on SQuAD\footnote{\url{https://github.com/google-research/bert}}. We fine-tune from the BERT-Base un-normalized multilingual model which includes Arabic. The model has 12-layers with $H=768$, 12-heads for self attention and inputs are padded to 384 tokens. We train on the training set of Arabic-SQuAD for 2 epochs with a learning rate of $3 \cdot 10^{-5}$. Similarly for QANet we modify the implementation of \footnote{\url{https://github.com/NLPLearn/QANet}} and use fastText embeddings and train for a total of 4 epochs. 

\textbf{Results and Analysis} We report all reader experiments in table \ref{table:reader}. The non-learning baselines are unable to obtain a significant improvement over a random guess on the EM and F1 metrics. The embedding and TF-IDF readers reach a sentence match accuracy of almost 75\%; this 75\% accuracy in fact corresponds to the percentage of word matching questions as in table \ref{table:questions}. On the other hand, BERT and QANet on the test set of Arabic-SQuAD reach 44.4 and 48.6 F1 scores respectively; as previously noted half of Arabic-SQuAD answers might be faulty as a result of NMT and this explains the relatively low results compared to the SQuAD leaderboard \cite{rajpurkar2016squad}. Now without having been trained on ARCD, both neural MRC models are able to perform well transferring knowledge from Arabic-SQuAD with BERT reaching a remarkable 90.08 SM accuracy.

\textbf{Transfer Learning.} To evaluate the effectiveness of using translated data as training data on the ARCD test set we train BERT under the following data regimes: (a) Arabic-SQuAD only, (b) ARCD-Train only and (c) Arabic-SQuAD and ARCD-Train combined; results are reported in table \ref{table:opensystem}. We remark that training under regimes (a) or (b) had very similar results, this gives strong evidence that Arabic-SQuAD could be in fact sufficient for obtaining powerful MRC models. When combining both datasets, we obtain an improvement of 8.3\% on the F1 score with a total score of 61.3; the training on ARCD allowed the model to better adapt to its differing answer distribution. 

\subsection{Open Domain QA}
We test our open domain approach SOQAL on ARCD-Test.
For our retriever we combine our hierarchical TF-IDF retriever with the Google Custom Search Engine to make sure we have a total of 10 retrieved articles. We train BERT on Arabic-SQuAD for two epochs and then fine-tune on ARCD-Train for an epoch.

We report in table \ref{table:opensystem} the accuracy of our proposed system on ARCD-Test achieving a 27.6 F1 and a 29.8 SM. The close F1 and SM scores indicate that the system is able to correctly retrieve the answer when it selects the correct paragraph, the issue then lies with it not scoring highly enough the correct paragraph. We also report the accuracy when the system outputs the top 3 and top 5 results (choosing the best answer out of them). 

\section{Conclusion}
To further the state of Arabic natural language understanding we proposed an approach for open domain Arabic QA and introduced the Arabic Reading Comprehension Dataset (ARCD) and Arabic-SQuAD: a machine translation of SQuAD \cite{devlin2018bert}. Our approach consisted of a document retriever using hierarchical TF-IDF and a document reader using BERT \cite{devlin2018bert}. We achieve a F1 score of 61.3 and a 90.0\% sentence match on ARCD and a 27.6 F1 score on an open domain version of ARCD. We also showed the effectiveness of using translated data as a training resource for QA.
Future work will aim to expand the size of ARCD and improve the end-to-end system by focusing on paragraph selection.

\bibliography{ref}
\bibliographystyle{acl_natbib}
\end{document}